\documentclass[letterpaper, 10 pt, conference]{ieeeconf}

\usepackage[hyphens]{url}
\usepackage{cite}
\usepackage{graphicx}
\usepackage{subcaption}
\usepackage{amsmath}
\usepackage{amssymb}
\usepackage{pgf}
\usepackage[acronym,nomain]{glossaries}
\usepackage{booktabs}

\newif\ifshowmargins
\showmarginsfalse

\ifshowmargins
  \usepackage{eso-pic}
  \usepackage{tikz}
  \AddToShipoutPictureBG{
    \begin{tikzpicture}[remember picture, overlay]
      \fill[gray!30] (current page.south west) rectangle ([xshift=0.75in]current page.north west);
      \fill[gray!30] ([xshift=-0.75in]current page.south east) rectangle (current page.north east);
      \ifnum\value{page}=1\relax
        \fill[gray!30] ([xshift=0.75in]current page.north west) rectangle ([xshift=-0.75in, yshift=-1in]current page.north east);
      \else
        \fill[gray!30] ([xshift=0.75in]current page.north west) rectangle ([xshift=-0.75in, yshift=-0.75in]current page.north east);
      \fi
      \fill[gray!30] ([xshift=0.75in]current page.south west) rectangle ([xshift=-0.75in, yshift=0.75in]current page.south east);
      \ifnum\value{page}=1\relax
        \node[font=\tiny, anchor=south west] at ([xshift=5pt, yshift=-15pt]current page.north west)
          {Margin requirements for first page};
        \node[font=\tiny, anchor=south west] at ([xshift=5pt, yshift=-25pt]current page.north west)
          {Paper size this page US Letter};
        \draw[<->] ([xshift=3.8in]current page.north west) -- ++(0, -1in)
          node[midway, right, font=\tiny] {72\,pt / 1\,in / 25.4\,mm};
      \fi
      \node[font=\tiny, rotate=90] at ([xshift=0.375in]current page.west) {54\,pt / 0.75\,in / 19.1\,mm};
      \node[font=\tiny, rotate=-90] at ([xshift=-0.375in]current page.east) {54\,pt / 0.75\,in / 19.1\,mm};
      \draw[<->] ([yshift=0.75in, xshift=3.8in]current page.south west) -- ++(0, -0.75in)
        node[midway, right, font=\tiny] {54\,pt / 0.75\,in / 19.1\,mm};
    \end{tikzpicture}
  }
\fi

\newacronym{amr}{AMR}{autonomous mobile robot}
\newacronym{mde}{MDE}{monocular depth estimation}
\newacronym{rl}{RL}{reinforcement learning}
\newacronym{drl}{DRL}{deep reinforcement learning}
\newacronym{ppo}{PPO}{Proximal Policy Optimization}
\newacronym{pomdp}{POMDP}{partially observable Markov Decision Process}
\newacronym{mse}{MSE}{mean squared error}
\newacronym{mlp}{MLP}{multi-layer perceptron}
\newacronym{fov}{FOV}{field of view}
\newacronym{tof}{ToF}{time-of-flight}
\newacronym{ros}{ROS}{Robot Operating System}
\newacronym{onnx}{ONNX}{Open Neural Network Exchange}
\newacronym{can}{CAN}{Controller Area Network}
\newacronym{vit}{ViT}{Vision Transformer}
\newacronym{kl}{KL}{Kullback-Leibler}
\newacronym{lidar}{LiDAR}{Light Detection and Ranging}

\title{\LARGE \bf
Learning Vision-Based Omnidirectional Navigation: A Teacher-Student Approach Using Monocular Depth Estimation
}

  \author{Jan Finke$^{1,2}$, Wayne Paul Martis$^{1,2}$, Adrian Schmelter$^{1,2}$,\\Lars Erbach$^{1,2}$, Christian Jestel$^{1,2}$, Marvin Wiedemann$^{1,2}$
  \thanks{This work was partially supported by Bundesministerium f{\"u}r Wirtschaft und Klimaschutz (BMWK), in the context of the project ``01ME23002F MediCar 4.0''.}
  \thanks{$^{1}$Fraunhofer Institute for Material Flow and Logistics, Dortmund, Germany}
  \thanks{$^{2}$The Lamarr Institute for Machine Learning and Artificial Intelligence, Germany}
  }

\begin{document}

\maketitle
\thispagestyle{empty}
\pagestyle{empty}

\begin{abstract}

Reliable obstacle avoidance in industrial settings demands 3D scene understanding, but widely used 2D LiDAR sensors perceive only a single horizontal slice of the environment, missing critical obstacles above or below the scan plane. We present a teacher-student framework for vision-based mobile robot navigation that eliminates the need for LiDAR sensors. A teacher policy trained via Proximal Policy Optimization (PPO) in NVIDIA Isaac Lab leverages privileged 2D LiDAR observations that account for the full robot footprint to learn robust navigation. The learned behavior is distilled into a student policy that relies solely on monocular depth maps predicted by a fine-tuned Depth Anything V2 model from four RGB cameras. The complete inference pipeline, comprising monocular depth estimation (MDE), policy execution, and motor control, runs entirely onboard an NVIDIA Jetson Orin AGX mounted on a DJI RoboMaster platform, requiring no external computation for inference. In simulation, the student achieves success rates of 82--96.5\%, consistently outperforming the standard 2D LiDAR teacher (50--89\%). In real-world experiments, the MDE-based student outperforms the 2D LiDAR teacher when navigating around obstacles with complex 3D geometries, such as overhanging structures and low-profile objects, that fall outside the single scan plane of a 2D LiDAR.
\end{abstract}

\glsresetall

\section{Introduction}

\Glspl{amr} are widely deployed in industrial environments to automate material transport between designated locations.
For navigation and collision avoidance, they predominantly rely on 2D \gls{lidar} sensors~\cite{LACKNER2024930}.
However, since this perception is limited to a single horizontal scan plane, \glspl{amr} struggle to operate reliably in complex scenarios involving obstacles that extend outside this plane, such as forklift tines, overhanging loads, or partially occupied shelves~\cite{Han2022}.
While supplementary 3D sensors such as stereo cameras or time-of-flight systems can mitigate this limitation~\cite{Gim2023, ada2023}, they add significant cost, weight, and integration complexity compared to standard RGB cameras~\cite{tosi2025}.

Recent advances in \gls{mde}, driven by large-scale foundation models, now yield highly accurate relative depth maps from single RGB images without dedicated depth sensors~\cite{Rajapaksha2024, Hu_2024}.
Foundation models such as Depth Anything V2~\cite{depth_anything_v2} can infer dense depth maps from single RGB images, effectively turning inexpensive off-the-shelf cameras into depth sensors while preserving rich spatial information.

\begin{figure}[t]
    \centering
    \vspace{0.25cm}
        \includegraphics[width=\columnwidth]{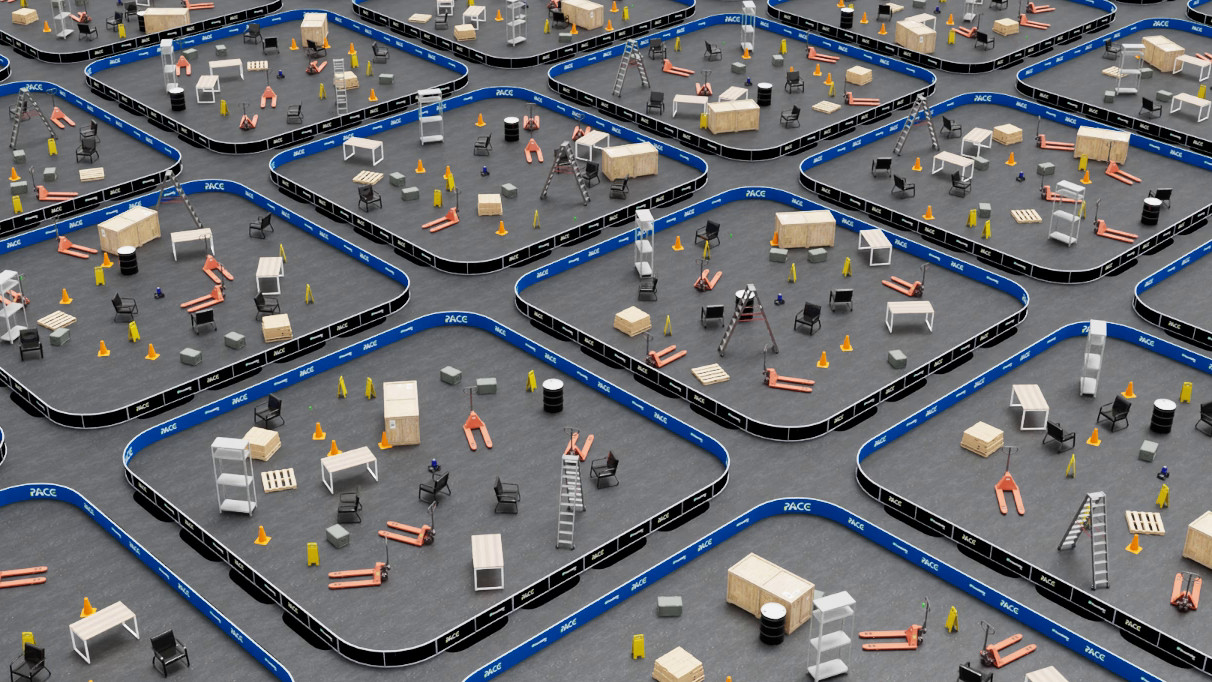}
    \caption{Parallel training environments in NVIDIA Isaac Lab. Each environment contains a randomized set of industrial obstacles for domain randomization during policy training.}
    \vspace{-0.2cm}
    \label{fig:parallel_envs}
\end{figure}

\begin{figure*}[t]
    \centering
        \includegraphics[width=\textwidth]{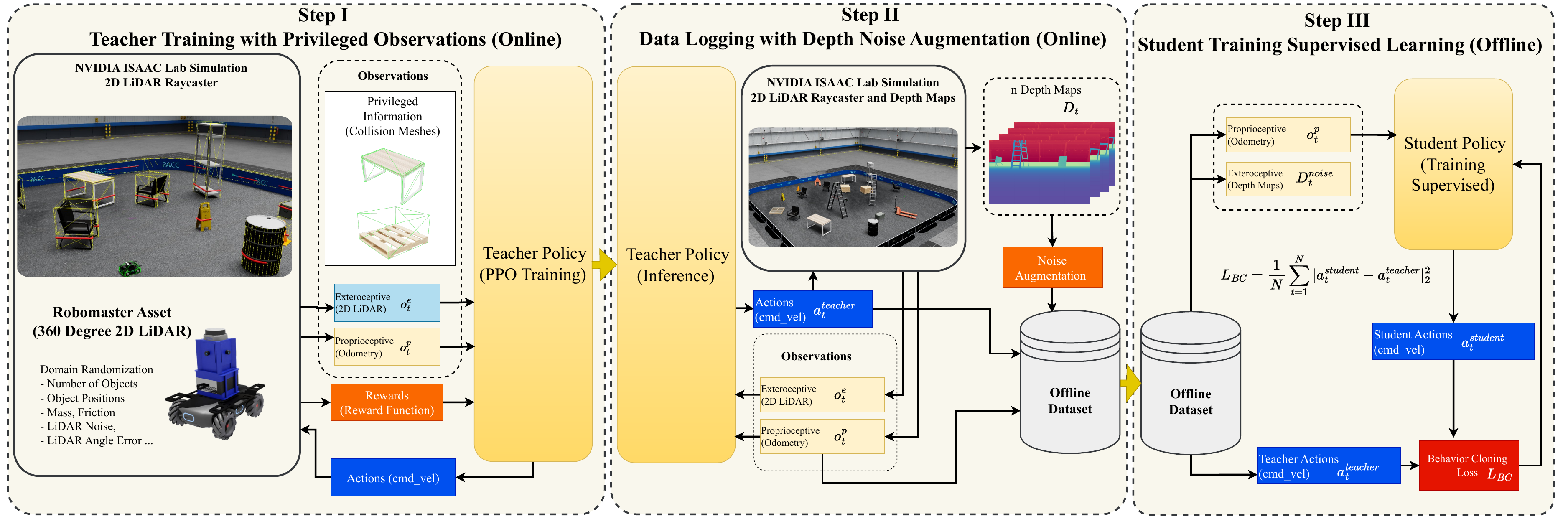}
    \caption{Teacher-student pipeline for learning vision-based navigation from privileged LiDAR observations.}
    \label{fig:system_overview}
\end{figure*}

In this work, we combine \gls{mde} with \gls{drl} to realize vision-based navigation, mapping depth observations directly to velocity commands without requiring hand-crafted planners or costmaps.
We adopt a teacher-student training scheme: a teacher policy is first trained with access to privileged collision information~\cite{chen2019learningcheating} in NVIDIA Isaac Lab~\cite{mittal2025isaaclab} (Fig.~\ref{fig:parallel_envs}).
The teacher's behavior is then distilled into a student policy that operates solely on depth maps obtained from four onboard RGB cameras through \gls{mde} at a control frequency of 10\,Hz, using a fine-tuned Depth Anything V2 model adapted to our domain.
Domain randomization~\cite{Tobin2017} and depth map augmentation~\cite{miki2022learning} facilitate robust sim-to-real transfer. An overview of the complete pipeline is shown in Fig.~\ref{fig:system_overview}.

Our main contributions are as follows:
\begin{itemize}
    \item A teacher-student \gls{rl} framework in which a \gls{ppo}-trained teacher with privileged 2D \gls{lidar} observations distills its policy into a student that navigates using only \gls{mde}-derived depth maps from a multi-camera setup.
    \item An onboard deployment on an NVIDIA Jetson Orin AGX embedded platform with successful sim-to-real transfer.
    \item A comparative evaluation in both simulation and real-world experiments demonstrating that the \gls{mde}-based student policy matches or exceeds the 2D \gls{lidar} teacher.
\end{itemize}

\section{Related Work}

Our work builds on three research areas: deep reinforcement learning for navigation, monocular depth estimation as a \gls{lidar} alternative, and privileged learning for policy distillation.

\subsection{Deep Reinforcement Learning for Mobile Robot Navigation}\label{sec:related_drl}

\Gls{drl} has emerged as a powerful paradigm for learning navigation policies that directly map sensory observations to control commands, bypassing the manual engineering required by traditional layered architectures~\cite{Tang2025}.
However, large-scale real-world evaluations reveal that the choice of sensor modality critically affects sim-to-real transfer~\cite{Gervet2023}.
Training such policies requires simulation environments that balance physical realism with computational efficiency~\cite{MuRoSim, mittal2025isaaclab}.
For training, \gls{ppo}~\cite{schulman2017ppo} has become the de facto standard for on-policy \gls{rl} due to its training stability and efficient scaling to massively parallel settings~\cite{rudin2022learning}.

\subsection{Monocular Depth Estimation for Mobile Robots}\label{sec:related_mde}

\Gls{lidar} sensors provide accurate 3D perception for mobile robot navigation but their high cost and power consumption~\cite{Gim2023} motivate the search for lightweight sensor alternatives in applications such as intralogistics~\cite{LACKNER2024930}.
While stereo cameras offer an alternative~\cite{tosi2025, ada2023}, monocular cameras provide the most lightweight solution, requiring only a single lens but depending on learned depth estimation~\cite{Dong2022, Ding2023}.
The advent of deep learning enabled end-to-end depth prediction from single images~\cite{Rajapaksha2024}, and recent foundation models achieve strong zero-shot generalization across diverse scenes.
These models can be categorized by output type: affine-invariant models such as Depth Anything V2~\cite{depth_anything_v2} predict relative depth with robust edge preservation, while metric models such as Metric3D v2~\cite{Hu_2024} recover absolute distances by modeling camera parameters.
Lightweight variants enable real-time inference on embedded platforms~\cite{Wofk2019, Feng2024}.
Since metrically perfect reconstruction is not required for navigation~\cite{Gao2021}, affine-invariant models are particularly suitable for learning-based approaches. Remaining challenges such as scale ambiguity and domain shift~\cite{Dong2022} motivate robust training strategies for sim-to-real transfer.

\subsection{Teacher-Student Learning and Sim-to-Real Transfer}\label{sec:related_ts}

Transferring policies from simulation to real robots remains a fundamental challenge. While domain randomization~\cite{Tobin2017} can partially bridge this gap, it becomes insufficient when deployed sensors differ fundamentally from those used during training.
The teacher-student paradigm addresses this by first training a teacher with access to privileged simulation-only information, then distilling its behavior into a student that uses only the sensor modality available during deployment~\cite{chen2019learningcheating}.
This approach has been successfully applied across diverse robotic domains including locomotion~\cite{miki2022learning}, planetary exploration~\cite{Mortensen2024}, and vision-based indoor navigation~\cite{Xu2025}.
A major constraint is computational cost: training \gls{rl} policies on depth maps requires rendering for thousands of environments, demanding substantial GPU memory~\cite{rudin2022learning} and limiting the parallelization critical for efficient training.

This motivates our approach: the teacher is trained on simulated laser scans with full parallelization, whereas the student is trained to navigate from simulated depth maps, with \gls{mde} applied during sim-to-real transfer.

\section{Methodology}

We propose a teacher-student framework that bridges \gls{mde} with \gls{rl} to enable vision-based robot navigation. The central challenge is the \emph{domain gap between rendered and \gls{mde}-predicted depth}: \gls{rl} policies trained on ground-truth depth cannot directly consume \gls{mde} outputs, which are noisy and partially accurate. Further, the computational overhead of simulating cameras hinders \gls{rl} policy training. Our framework resolves this through a teacher-student training pipeline. An expert teacher policy trained with privileged 2D \gls{lidar} achieves robust navigation via parallel simulation~\cite{rudin2022learning}, then distills its behavior offline into a student that operates on noise-augmented depth maps. Domain randomization~\cite{Tobin2017} and depth augmentation facilitate sim-to-real transfer. Section~\ref{sec:problem_formulation} formalizes the navigation problem. Section~\ref{sec:mde} describes the \gls{mde} module. Sections~\ref{sec:teacher_policy} and~\ref{sec:student_policy} detail the teacher and student policy training, respectively.

\subsection{Problem Formulation}\label{sec:problem_formulation}

We address collision-free goal-directed navigation for an omnidirectional mobile robot in indoor environments with static obstacles, formulated as a \gls{pomdp} $(\mathcal{S}, \mathcal{A}, \mathcal{O}, R, P, \gamma)$ with state space $\mathcal{S}$, action space $\mathcal{A}$, observation space $\mathcal{O}$, reward function $R$, transition dynamics $P$, and discount factor $\gamma \in [0,1)$.
The robot receives partial observations $o_t \in \mathcal{O}$ of the true state $s_t \in \mathcal{S}$ and selects actions $a_t \in \mathcal{A}$ according to a policy $\pi_\theta(a_t \mid o_t)$.
Crucially, the observation space differs between teacher and student: the teacher observes privileged 2D \gls{lidar} scans (Section~\ref{sec:teacher_policy}), while the student receives monocular depth maps (Section~\ref{sec:student_policy}). Both policies share the action space and proprioceptive observations, but only the teacher requires the reward function, as the student is trained offline via behavior cloning.

\subsection{Monocular Depth Estimation Backbone}\label{sec:mde}
As established in Section~\ref{sec:related_mde}, affine-invariant depth models are sufficient for learning-based navigation. We select Depth Anything V2~\cite{depth_anything_v2} in its \gls{vit}-Small variant (25M parameters) for its zero-shot generalization and compatibility with embedded inference on NVIDIA Jetson platforms.
To reduce the domain gap between rendered and \gls{mde}-predicted depth, we fine-tune the model on domain-specific metric data from a warehouse environment.

\begin{figure}[t]
    \centering
    \includegraphics[width=\columnwidth]{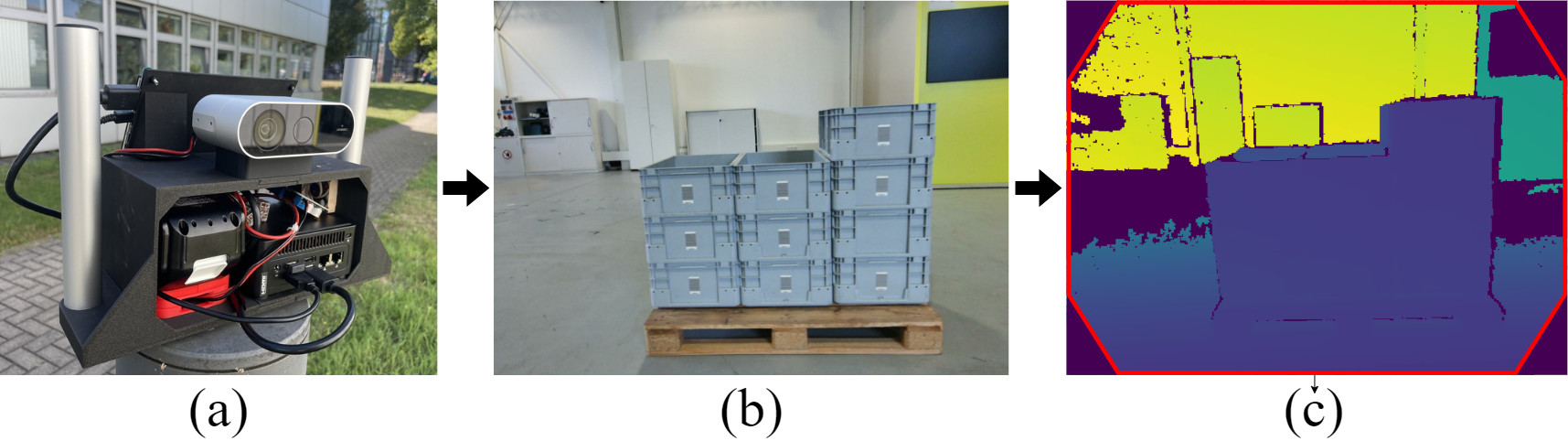}
    \caption{Data collection setup for fine-tuning the MDE backbone: (a)~recording platform with Orbbec Femto Bolt ToF camera, (b)~example RGB image, and (c)~corresponding ground-truth depth map.}
    \label{fig:rec_platform}
\end{figure}

\textbf{Fine-Tuning of Depth Anything V2:}
For fine-tuning, we recorded a domain-specific dataset of 24,499 RGB and depth map pairs using an Orbbec Femto Bolt \gls{tof} camera (Fig.~\ref{fig:rec_platform}), covering typical logistics scenarios such as loaded pallets, industrial trucks, and high-bay warehouses. The \gls{vit}-Small model from Depth Anything V2 was then fine-tuned for 80 epochs on the training split (77.6\%).

Fine-tuning resulted in an improvement of 118.4\% in the $\delta_1$ accuracy on the test data split compared to the metric \gls{vit}-Small model from Depth Anything V2, which was fine-tuned exclusively on the Hypersim dataset~\cite{hypersim}. Here, $\delta_1$ denotes the fraction of pixels satisfying $\max(d^*/d,\; d/d^*) < 1.25$, where $d$ and $d^*$ are the predicted and ground-truth depth, respectively.

\textbf{Deployment:}
We export the fine-tuned model to TensorRT (BF16) and benchmark batch inference of four camera images at $480 \times 300$\,pixels on three NVIDIA Jetson platforms (Table~\ref{tab:mde_inference}). We select the Jetson Orin AGX for deployment, which completes a batch in $50.16 \pm 8.22$\,ms ($19.9$\,FPS), providing sufficient headroom for the 10\,Hz control frequency required by the navigation policy.

\begin{table}[t]
    \centering
    \caption{Inference performance of the fine-tuned \gls{mde} model on NVIDIA Jetson platforms.}
    \label{tab:mde_inference}
    \begin{tabular}{lccc}
        \toprule
        \textbf{Platform} & \textbf{JetPack} & \textbf{Latency [ms]} & \textbf{Avg. FPS} \\
        \midrule
        NVIDIA Jetson Orin NX  & 6.1 & $98.63 \pm 2.06$  & 10.1 \\
        NVIDIA Jetson Orin AGX & 6.1 & $50.16 \pm 8.22$ & 19.9 \\
        NVIDIA Jetson Thor     & 7.0 & $19.02 \pm 4.81$ & 52.6 \\
        \bottomrule
    \end{tabular}
\end{table}

\subsection{Teacher Policy}\label{sec:teacher_policy}
The teacher policy is trained with access to privileged simulation-only information, enabling it to learn a robust navigation behavior that the student later imitates using only onboard sensors. The following paragraphs detail the \gls{pomdp} formulation and training procedure.

\begin{figure}[t]
    \centering
    \includegraphics[width=\columnwidth]{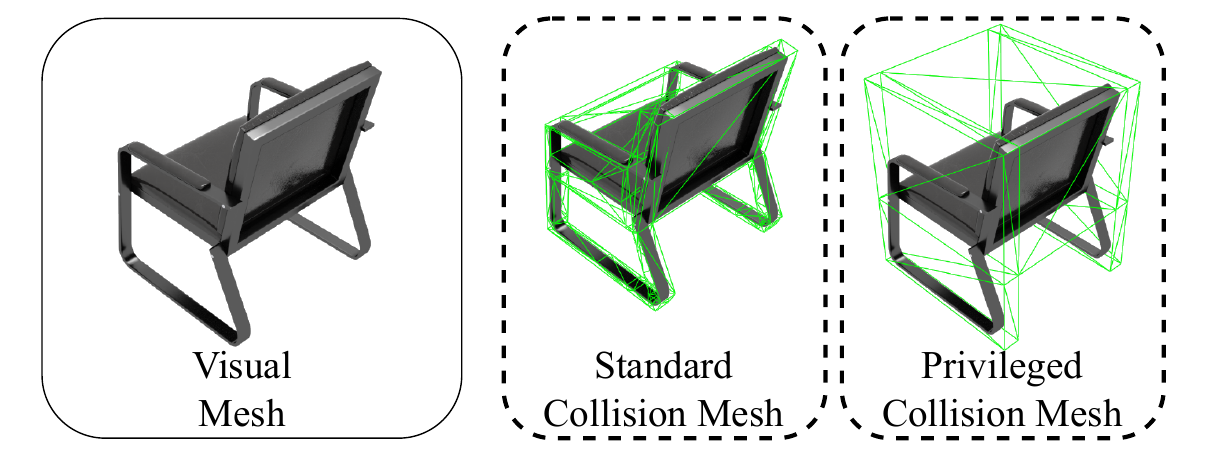}
    \caption{Collision mesh representation used for privileged \gls{lidar}-based observations in the teacher training phase.}
    \label{fig:colliders}
\end{figure}

\textbf{Observation Space:}
The teacher policy receives privileged observations composed of exteroceptive and proprioceptive inputs:
\begin{equation}
    o_t^{\text{teacher}} = (o_t^e,\; o_t^p)
\end{equation}
The exteroceptive input is a 2D \gls{lidar} scan with $n_{\text{scan}} = 3000$ range measurements:
\begin{equation}
    o_t^e = l_t \in \mathbb{R}^{n_{\text{scan}}}
\end{equation}
The proprioceptive input combines odometry, goal information, and the previous action:
\begin{equation}
    o_t^p = (s_t^{\text{odom}},\; \mathbf{p}_{\text{robot}}^{\text{goal}},\; d_{\text{goal}},\; a_{t-1})
\end{equation}
The odometry history stores the current and past $n_{\text{odom}} = 5$ velocity estimates $v^{\text{odom}} = (v_x, v_y, \omega_z)^T$:
\begin{equation}
    s_t^{\text{odom}} = (v_t^{\text{odom}},\; v_{t-1}^{\text{odom}},\; \dots,\; v_{t-n_{\text{odom}}}^{\text{odom}})
\end{equation}
This history enables the policy to implicitly learn the robot's dynamics. The unit vector $\mathbf{p}_{\text{robot}}^{\text{goal}}$ points toward the goal in the robot's coordinate frame, and $d_{\text{goal}}$ denotes the Euclidean goal distance. Notably, the relative angle to the goal $\theta_{\text{goal}}$ is omitted, as the robot's omnidirectional mobility makes orientation-independent navigation preferable.
In simulation, the \gls{lidar} observations are obtained via raycasting against privileged collision meshes that extend beyond the true object boundaries, preventing the robot from entering gaps narrower than its own footprint. An example of an object with a privileged collider is shown in Fig.~\ref{fig:colliders}. The resulting distance measurements inherently block regions where the robot cannot physically fit, so the teacher only perceives passages that are actually traversable. This privileged information is unavailable during real-world deployment and is therefore not used by the student policy.

\begin{samepage}
\textbf{Action Space:}
The policy outputs normalized actions $a_t^{\text{teacher}} = (v_x, v_y, \omega_z) \in [-1, 1]^3$ via a $\tanh$ activation, which are scaled to physical velocity commands with $v_{x,\max} = v_{y,\max} = 0.5$\,m/s and $\omega_{z,\max} = 1.0$\,rad/s.
Actions are sent to the robot at a control frequency of 10\,Hz.
\end{samepage}

\textbf{Reward Function:}
The reward at each timestep $t$ is a piecewise function combining sparse terminal rewards and weighted continuous shaping terms:
\begin{equation}
    r_t = \begin{cases}
        r_{\text{finished}}, & \text{if goal reached} \\
        r_{\text{collision}}, & \text{if collision occurred} \\
        0, & \text{if timeout} \\
        r_{\text{at\_goal}} + \sum_i w_i r_i^t, & \text{otherwise}
    \end{cases}
\end{equation}
where $r_{\text{finished}}$ is a large positive sparse reward for successful goal reaching, $r_{\text{collision}}$ a corresponding penalty, and $r_{\text{at\_goal}}$ a discrete bonus triggered once upon entering the goal radius $d_{\text{goal}}^{\text{thresh}}$.
The continuous shaping terms are:
\begin{itemize}
    \item $r_{\Delta d}^t$: delta distance reward for progress toward the goal,
    \item $r_{\text{lidar}}^t$: minimum \gls{lidar} distance penalty discouraging proximity to obstacles within $d_{\text{min}}$,
    \item $r_a^t$: action magnitude penalty,
    \item $r_{\Delta a}^t$: action rate penalty to smooth velocity commands.
\end{itemize}
The weights $(w_{\Delta d}, w_{\text{lidar}}, w_a, w_{\Delta a})$ control the relative importance of each component. All reward parameters are listed in Table~\ref{tab:reward_weights}.
Episodes terminate upon goal reaching, collision, or timeout.

\begin{figure}[t]
    \centering
    \includegraphics[width=\columnwidth, trim=0.5cm 0 0.5cm 0, clip]{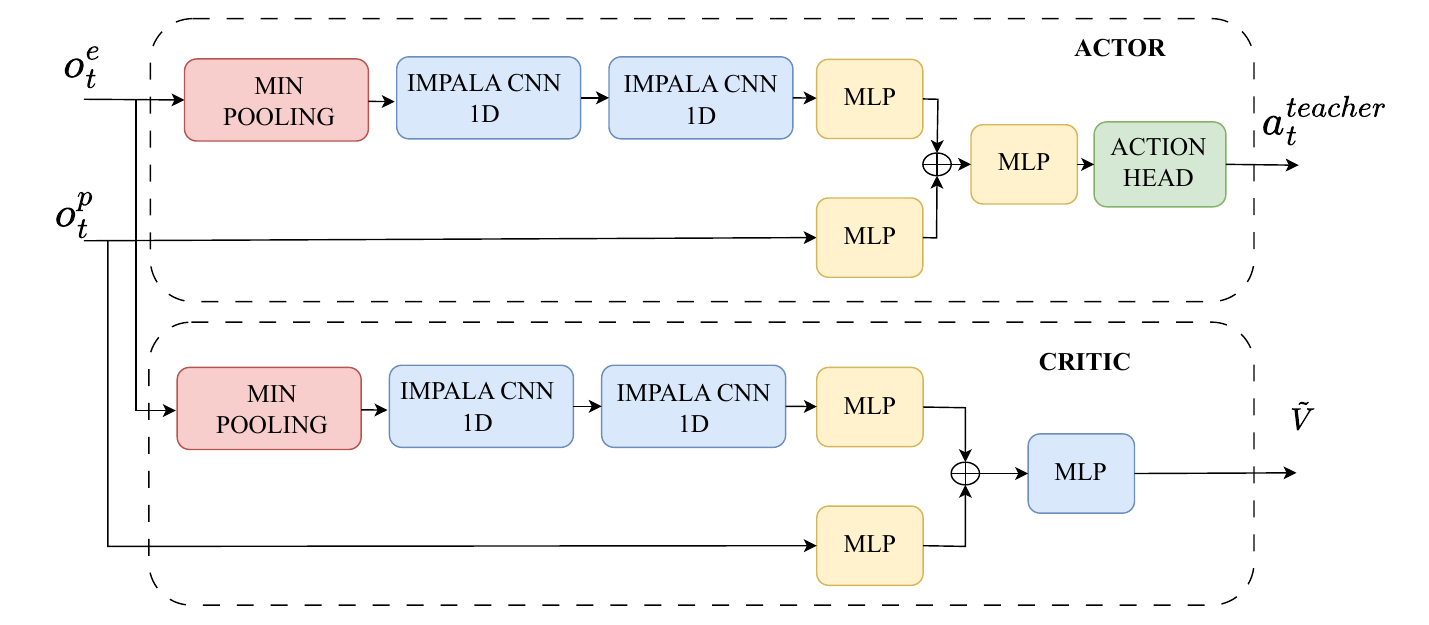}
    \caption{Teacher policy network architecture with separate actor and critic. Both process exteroceptive \gls{lidar} observations $o_t^e$ via an IMPALA encoder and proprioceptive input $o_t^p$ via an \gls{mlp}.}
    \label{fig:teacher_network}
\end{figure}

\textbf{Network Architecture:}
The teacher policy uses an actor-critic architecture, as illustrated in Fig.~\ref{fig:teacher_network}. Both networks share a nearly identical structure, processing exteroceptive and proprioceptive inputs through separate encoding pathways.
The exteroceptive \gls{lidar} input $o_t^e$ is first downsampled via min-pooling (kernel size 10) to retain nearest-obstacle information, then encoded by two consecutive 1D IMPALA-CNN blocks~\cite{espeholt2018impala} with 24 and 16 filters, respectively. The proprioceptive input $o_t^p$ is processed independently through a 96-unit linear layer with ReLU activation.
Both representations are concatenated and passed through a shared 256-unit \gls{mlp}. The actor's output head applies a $\tanh$ activation to bound the action output to $[-1, 1]$, while the critic produces a scalar value estimate for advantage computation.

\begin{table}[t]
    \centering
    \caption{Reward function parameters and weights.}
    \label{tab:reward_weights}
    \begin{tabular}{lrlr}
        \toprule
        \textbf{Parameter} & \textbf{Value} & \textbf{Parameter} & \textbf{Value} \\
        \midrule
        $w_{\Delta d}$ & 0.5 & $w_{\Delta a}$ & -0.1 \\
        $w_{\text{lidar}}$ & -0.1 & $r_{\text{at\_goal}}^t$ & 10.0 \\
        $d_{\text{min}}$ & 0.7 m & $d_{\text{goal}}^{\text{thresh}}$ & 0.5 m \\
        $w_a$ & -0.05 & $r_{\text{collision}}$ & -50.0 \\
        $r_{\text{finished}}$ & 50.0 & & \\
        \bottomrule
    \end{tabular}
\end{table}

\textbf{Training:}
The teacher policy is trained online using the \gls{ppo}~\cite{schulman2017ppo} implementation provided by Isaac Lab~\cite{mittal2025isaaclab}.
Training runs across 1024 parallel environments at a simulation rate of 360\,Hz with a control frequency of 10\,Hz.
The Adam optimizer is used for gradient updates with an adaptive learning rate schedule based on \gls{kl} divergence, which measures the deviation between successive policy distributions and reduces the learning rate when updates become too large. Rollouts of horizon length $T = 96$ are collected before each update step.
Key \gls{ppo} hyperparameters are listed in Table~\ref{tab:ppo_hyperparams}.

\begin{table}[t]
    \centering
    \caption{\gls{ppo} hyperparameters for teacher policy training.}
    \label{tab:ppo_hyperparams}
    \begin{tabular}{lrlr}
        \toprule
        \textbf{Parameter} & \textbf{Value} & \textbf{Parameter} & \textbf{Value} \\
        \midrule
        Learning rate ($\alpha$) & $1 \times 10^{-3}$ & Discount factor ($\gamma$) & 0.99 \\
        Horizon ($T$) & 96 & Clip range ($\epsilon$) & 0.2 \\
        Mini-epochs & 5 & Batch size & 98304 \\
        Minibatch size & 24576 & Environments ($N_{\text{env}}$) & 1024 \\
        \bottomrule
    \end{tabular}
\end{table}

\subsection{Student Policy}\label{sec:student_policy}
The following paragraphs describe the data collection from the teacher, the noise augmentation pipeline for bridging the sim-to-real gap, and the student's network architecture and training procedure.

\textbf{Data Logging:}
Expert demonstrations are collected by executing the trained teacher policy in inference mode within the Isaac Lab simulation.
At each timestep $t$, the teacher produces expert actions $a_t^{\text{teacher}}$ from its \gls{lidar} observations. Simultaneously, four depth maps $D_t$ and the proprioceptive state $o_t^p$ are recorded alongside the teacher's actions, yielding tuples $(D_t, o_t^p, a_t^{\text{teacher}})$.
Domain randomization across episodes ensures diverse navigation scenarios.
The dataset consists of 28,838 successful episodes, with collision episodes filtered out to ensure high-quality expert demonstrations.

\textbf{Noise Augmentation:}
To bridge the sim-to-real gap between perfect rendered depth and real \gls{mde} outputs, we apply a seven-stage augmentation pipeline to simulated depth maps during data collection, inspired by~\cite{miki2022learning,Mortensen2024}.
The pipeline includes: (1) Gaussian smoothing for lens blur, (2) motion blur artifacts, (3) background smudging for far-range degradation, (4) elastic smearing for edge distortions, (5) random scale-shift perturbations to model \gls{mde} affine-invariance~\cite{depth_anything_v2}, (6) low-frequency Gaussian noise for environmental effects, and (7) quantization noise simulating compression.

\begin{figure}[t]
    \centering
    \vspace{0.2cm}
    \hspace*{-0.1cm}
    \includegraphics[width=1.0\columnwidth, trim=0 0 1.9cm 0, clip]{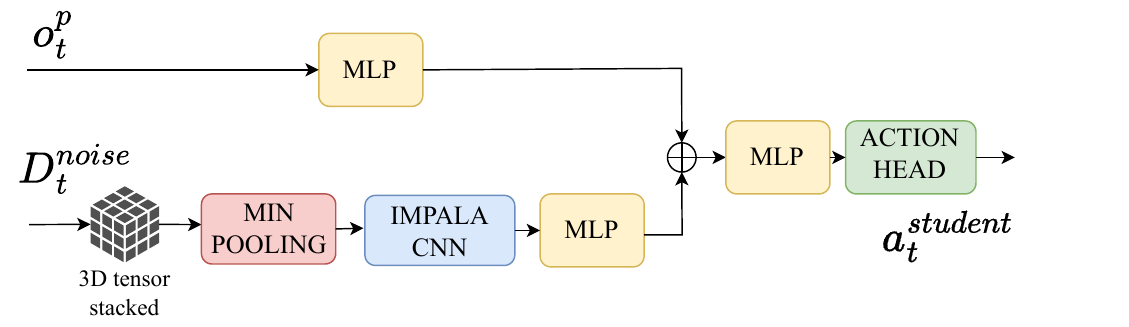}
    \caption{Student network architecture. It is similar to the teacher actor network but instead of 2D \gls{lidar} data, it processes 4 depth maps through an IMPALA-CNN encoder followed by an \gls{mlp}.}
    \label{fig:student_network}
\end{figure}

\textbf{Network Architecture:}
The student policy shares several architectural components with the teacher, as illustrated in Fig.~\ref{fig:student_network}.
Key hyperparameters such as the \gls{mlp} width (256 and 96 units) are shared between teacher and student to maintain comparable latent representations.
The student operates on visual observations $o_t^{\text{student}} = (D_t, o_t^p)$, where $D_t \in \mathbb{R}^{4 \times 300 \times 480}$ is a stack of four depth maps from four cameras covering a $360^{\circ}$ \gls{fov}, replacing the privileged \gls{lidar} input. The proprioceptive observations $o_t^p$ remain identical to those of the teacher.
The key distinction from the teacher lies in the exteroceptive pathway.
The four depth maps are stacked along their channel dimension to form a 3D tensor, which is first downsampled via min-pooling with a $5 \times 5$ kernel to preserve nearest-obstacle information.
The downsampled tensor is then passed through a 2D IMPALA-CNN~\cite{espeholt2018impala} encoder with three sequential blocks (feature dimensions $[16, 32, 32]$), followed by a ReLU activation and a fully connected layer with 256 hidden units.
In parallel, the proprioceptive input $o_t^p$ is processed through a 96-unit \gls{mlp}, consistent with the teacher network.
Both latent representations are concatenated and forwarded through a shared 256-unit \gls{mlp} to produce the final action output, bounded to $[-1, 1]$ via $\tanh(\cdot)$ activation.

\textbf{Training:}
The student policy is trained offline via supervised imitation learning (behavior cloning) on the logged dataset.
At each timestep, the student receives noise-augmented depth maps $D_t^{\text{noise}}$ and proprioceptive observations $o_t^p$ as input, and is expected to reproduce the expert actions $a_t^{\text{teacher}}$ of the teacher.
The noise augmentation simulates the degradation modalities of real \gls{mde} outputs encountered during deployment, thereby exposing the student to a representative distribution of depth artifacts and enabling robust imitation of the teacher's expert behavior under realistic sensing conditions.
Depth maps are clipped to a maximum range of 5\,m to suppress irrelevant background information.

Training minimizes the behavior cloning loss, defined as the \gls{mse} between the teacher's ground-truth actions and the student's predicted actions:
\begin{equation}
    \mathcal{L}_{\text{BC}} = \frac{1}{N} \sum_{i=1}^{N} \left\| a_i^{\text{teacher}} - a_i^{\text{student}} \right\|^2
\end{equation}
where $N$ is the number of samples in a batch, $a_i^{\text{teacher}}$ is the expert action, and $a_i^{\text{student}} = \pi_\theta(D_i^{\text{noise}}, o_i^p)$ is the student's predicted action for the $i$-th sample.
The student is optimized using the Adam optimizer with a constant learning rate of $5 \times 10^{-4}$, a batch size of 8, and trained for 100 epochs.

\section{Experiments}

We evaluate our framework in simulation and on the real robot platform.

\subsection{Isaac Lab Simulation Environment}

Training is conducted on a single NVIDIA RTX~4090 GPU with 24\,GB VRAM. The teacher policy is trained in Isaac Sim~5.1 with the Isaac Lab framework, running 1024 parallel environments for large-scale \gls{rl} (Fig.~\ref{fig:parallel_envs}). The training uses three arena configurations of varying size to encourage generalization across different spatial scales. In each episode, obstacles sampled from a pool of 15 distinct object types are randomly placed within the arena to create diverse navigation scenarios. The student policy is subsequently trained offline on the same GPU using the logged demonstration data from the teacher (see Section~\ref{sec:student_policy}).

\subsection{Hardware Platform and Camera Calibration}\label{sec:hardware_calibration}

For our experiments, we utilize the DJI RoboMaster S1, which features four Mecanum wheels for omnidirectional movement.
The robot is equipped with an NVIDIA Jetson Orin AGX, which is used for onboard inference of the student policy and to run the \gls{mde} model locally.
To have a full view of the robot's surroundings, the robot is equipped with four RGB cameras with AR0234 global shutter image sensors. These sensors are connected to a Luxonis FFC4 module, which generates a hardware trigger to synchronize all cameras and outputs images at a resolution of $480 \times 300$\,pixels.
The sensors are equipped with Arducam LN056 wide-angle lenses. All cameras are calibrated to each other using the Basalt framework~\cite{Usenko2019VisualInertialMW} and an AprilTag~\cite{AprilTag} calibration target.
To undistort the cameras, we upload the Kannala-Brandt calibration parameters~\cite{KannalaBrandt} to the Luxonis boards, yielding a \gls{fov} of $93.97^{\circ} \times 67.64^{\circ}$ for each camera.

Additionally, an RPLIDAR S2 2D \gls{lidar} sensor is mounted on top of the robot, providing $360^{\circ}$ range measurements. Because the angular offset of the raw scan varies between sweeps, the measurements are rebinned into $n_{\text{scan}} = 3000$ fixed angular positions to match the simulation raycasting layout. This enables deployment of the teacher policy on the real robot for baseline comparison with the \gls{mde}-based student policy.

\subsection{Simulation Asset}
\begin{figure}[t]
    \centering
    \vspace{0.2cm}
        \includegraphics[width=\columnwidth]{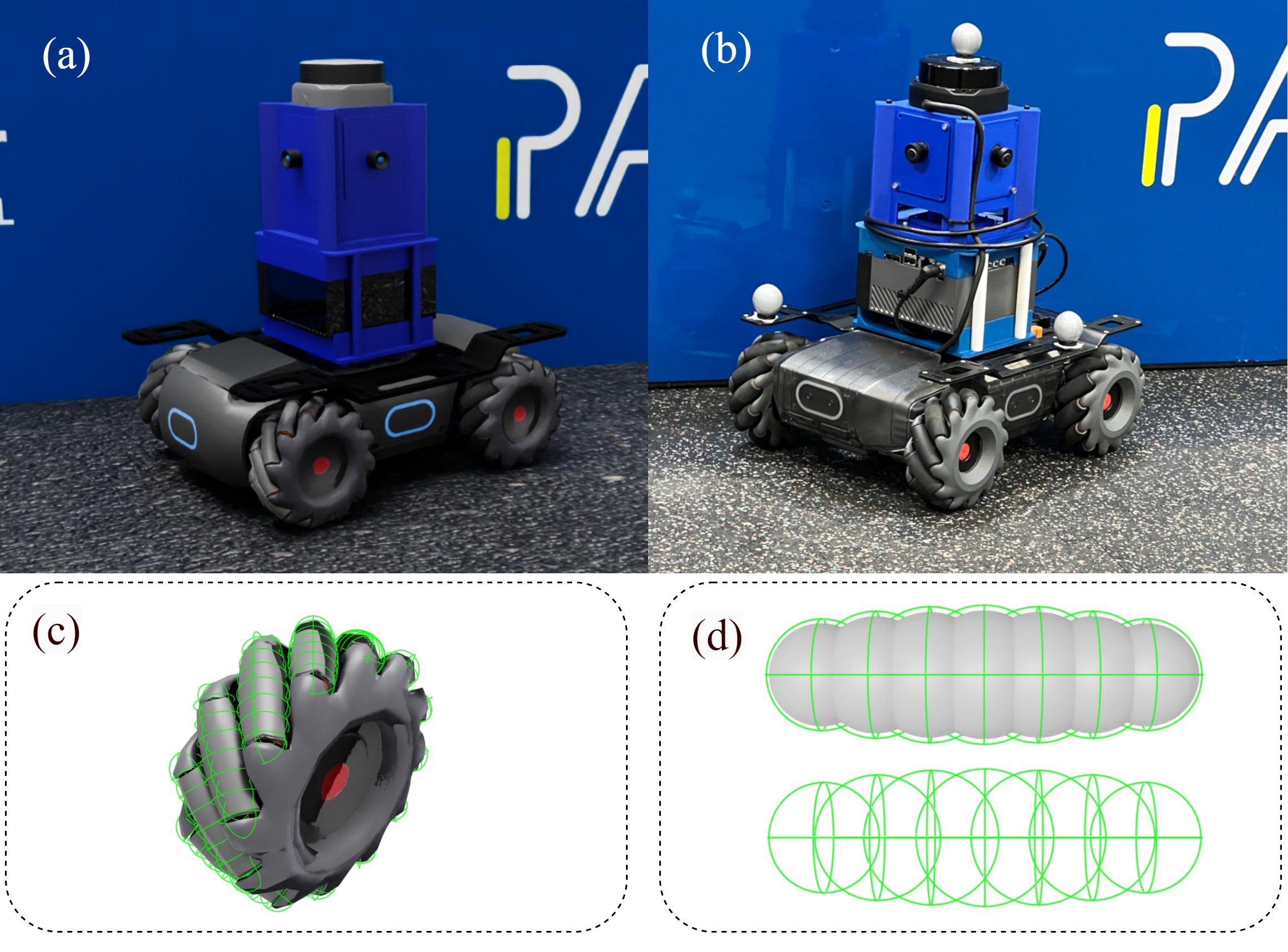}
    \caption{DJI RoboMaster platform: (a)~simulation model, (b)~real robot, (c)~Mecanum wheel with sphere colliders, and (d)~detailed collider geometry of a single Mecanum roller.}
    \label{fig:robomaster_asset}
\end{figure}

The simulation asset is shown in Fig.~\ref{fig:robomaster_asset}.
The robot model consists of four Mecanum wheels, each with 12 free-rolling rollers. Following the approach of~\cite{Wiedemann2024}, each roller is approximated by 7 sphere colliders. The physical parameters of the asset, such as friction and joint damping, are optimized using recorded ground-truth motion data and the system identification approach from~\cite{Wiedemann2024,Wiedemann2025}. Additionally, a deadzone is applied to each wheel to account for the inability of the RoboMaster's motors to actuate at low velocities.

\subsection{Simulation Results: Teacher Policy}

\begin{figure}[t]
    \centering
    \resizebox{0.92\columnwidth}{!}{\input{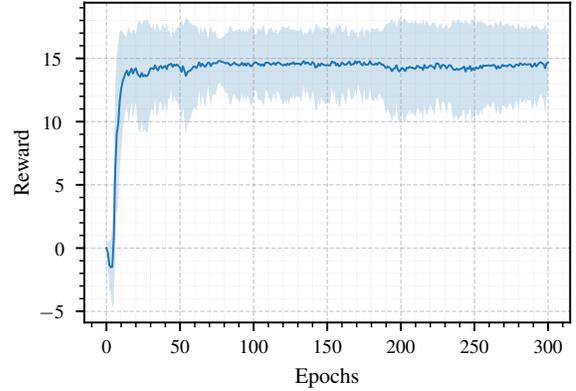}}
    \caption{Training reward progression during teacher policy training.}
    \label{fig:training_rewards}
\end{figure}

\begin{figure}[t]
    \centering
    \includegraphics[width=0.88\columnwidth]{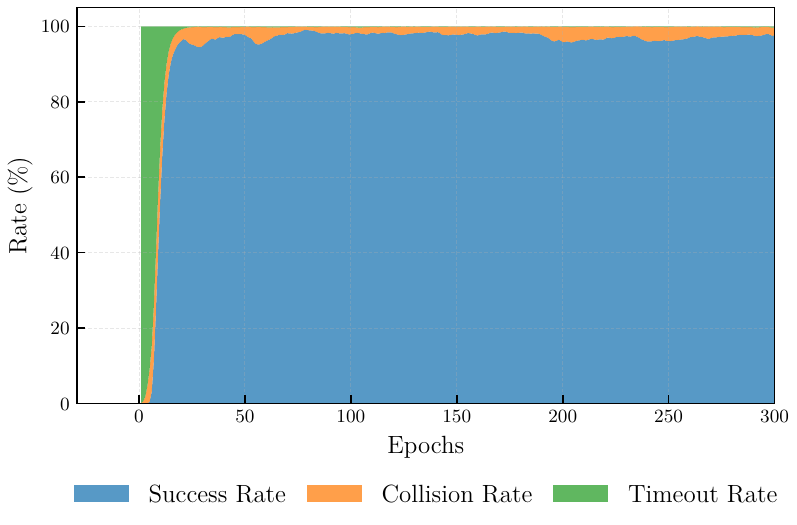}
    \caption{Success rate distribution of the teacher policy across evaluation episodes.}
    \label{fig:total_rate_box}
\end{figure}

The teacher policy is trained for 300 epochs, completing in approximately 120 minutes. Fig.~\ref{fig:training_rewards} shows the average episode reward over training. The reward increases steeply within the first 25 epochs and plateaus at approximately 14.5 after epoch 50, indicating fast and stable convergence of the teacher policy. Fig.~\ref{fig:total_rate_box} illustrates the episode termination distribution: the policy transitions from timeouts through a collision-heavy phase to mostly successful episodes, reaching a peak success rate of 98.5\%, noting that training episodes sample a higher and variable number of objects than the fixed evaluation configurations in Table~\ref{tab:simulation_results}.

\subsection{Simulation Results: Student Policy}

Table~\ref{tab:simulation_results} summarizes the success rates across simulation environments with varying obstacle density. In these experiments, the teacher operates with standard (non-privileged) collision meshes, removing the traversability advantage it had during training. Under these conditions, the depth-based student consistently outperforms the \gls{lidar}-based teacher. Without the privileged collision meshes, the teacher's 2D \gls{lidar} scans miss obstacles outside the scan plane, leading to a steep performance drop as obstacle density increases. The student's depth maps, by contrast, capture the full vertical extent of obstacles, maintaining higher success rates across all configurations.
\begin{table}[t]
    \centering
    \caption{Teacher and student success rates in simulation for varying obstacle counts. Std.\ and Priv.\ refer to standard and privileged collision meshes, respectively (see Fig.~\ref{fig:colliders}).}
    \label{tab:simulation_results}
    \begin{tabular}{cccc}
        \toprule
        \textbf{Obstacle} & \textbf{Teacher} & \textbf{Teacher} & \textbf{Student} \\
        \textbf{Counts} & \textbf{\gls{lidar} (Std.)} & \textbf{\gls{lidar} (Priv.)} & \textbf{MDE (Std.)} \\
        \midrule
        10 & 89\%   & 99.5\% & 96.5\% \\
        15 & 73.5\% & 93\%   & 94.5\% \\
        20 & 58\%   & 95\%   & 92\%   \\
        25 & 50\%   & 89.5\% & 82\%   \\
        \bottomrule
    \end{tabular}
\end{table}

\subsection{Real-World Results}

\begin{figure}[t]
    \centering
    \vspace{0.1cm}
        \includegraphics[width=\columnwidth]{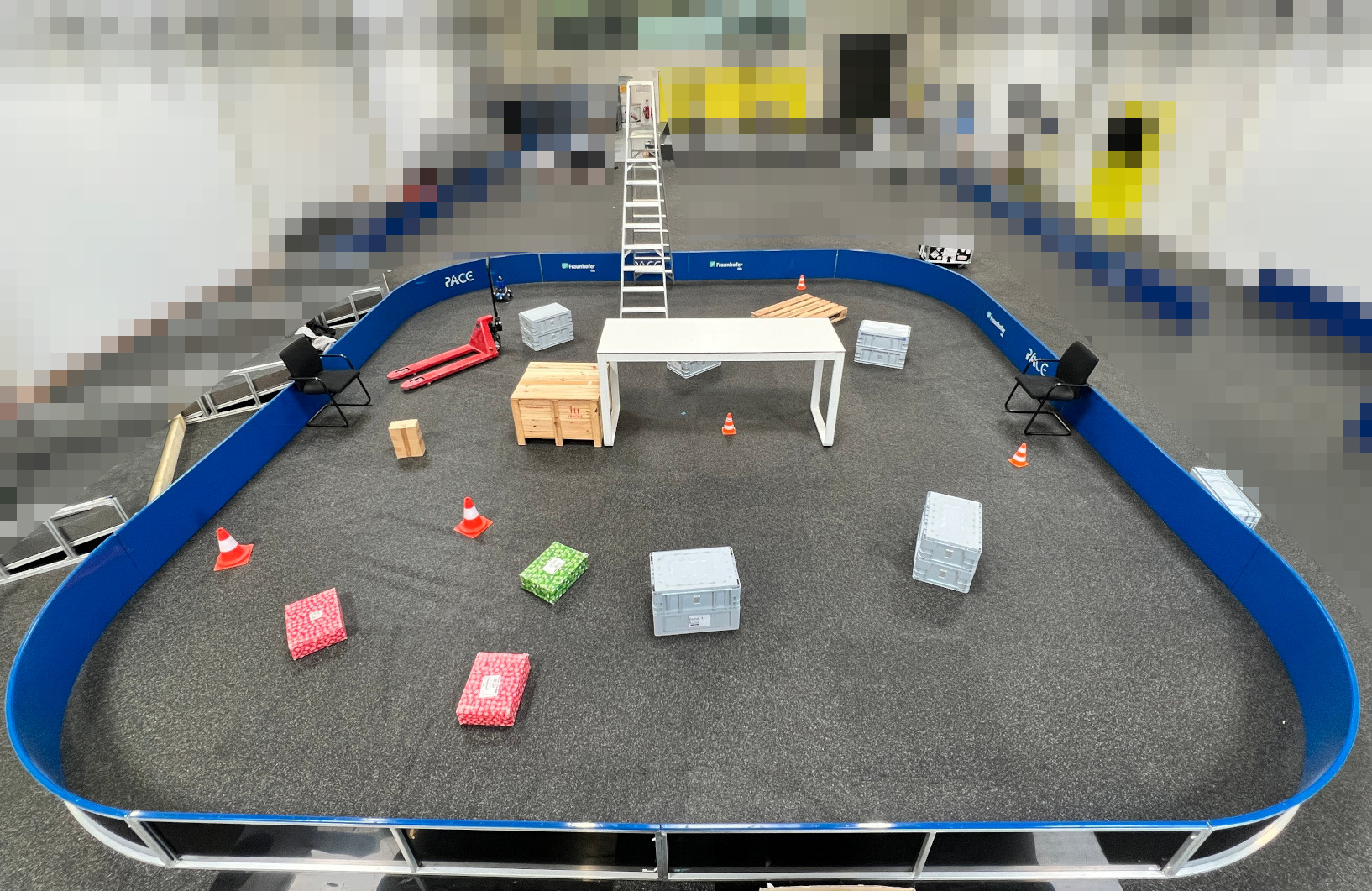}
    \caption{Real-world test environment: an 8\,m\,$\times$\,8\,m arena with obstacles of varying size and geometry used for sim-to-real evaluation.}
    \label{fig:sim2real}
\end{figure}

To evaluate sim-to-real transfer, we deploy both policies on the real robot in an 8\,m\,$\times$\,8\,m arena containing objects of varying size, as shown in Fig.~\ref{fig:sim2real}. Robot trajectories are recorded using a VICON motion capture system, which provides ground-truth goal distance and direction to both policies, isolating the exteroceptive perception (2D \gls{lidar} vs.\ \gls{mde}) as the sole variable under evaluation.

For this scenario, we measured the success rate of the teacher and student policies deployed on the RoboMaster for three different endpoints from a common starting point, with 10 runs per endpoint.

\begin{table}[t]
    \centering
    \caption{Real-world success rate for teacher and student across three endpoints with 10 runs each.}
    \label{tab:real_world_results}
    \begin{tabular}{lcc}
        \toprule
        \textbf{Scenario} & \textbf{Teacher} & \textbf{Student} \\
         & \textbf{\gls{lidar}} & \textbf{\gls{mde}} \\
        \midrule
        Endpoint A & 40\% & 80\% \\
        Endpoint B & 30\% & 60\% \\
        Endpoint C & 40\% & 100\% \\
        \bottomrule
    \end{tabular}
\end{table}

Table~\ref{tab:real_world_results} shows the success rates of the real-world experiments. The \gls{mde}-based student consistently outperforms the \gls{lidar}-based teacher across all three endpoints, achieving an average success rate of 80\% compared to 37\% for the teacher. The teacher's failures are predominantly caused by collisions with low-lying obstacles outside the \gls{lidar} scan plane, which the depth-based student perceives through its full-frame \gls{fov}.

\begin{figure}[t]
    \centering
    \vspace{-0.05cm}
    \includegraphics[width=\columnwidth]{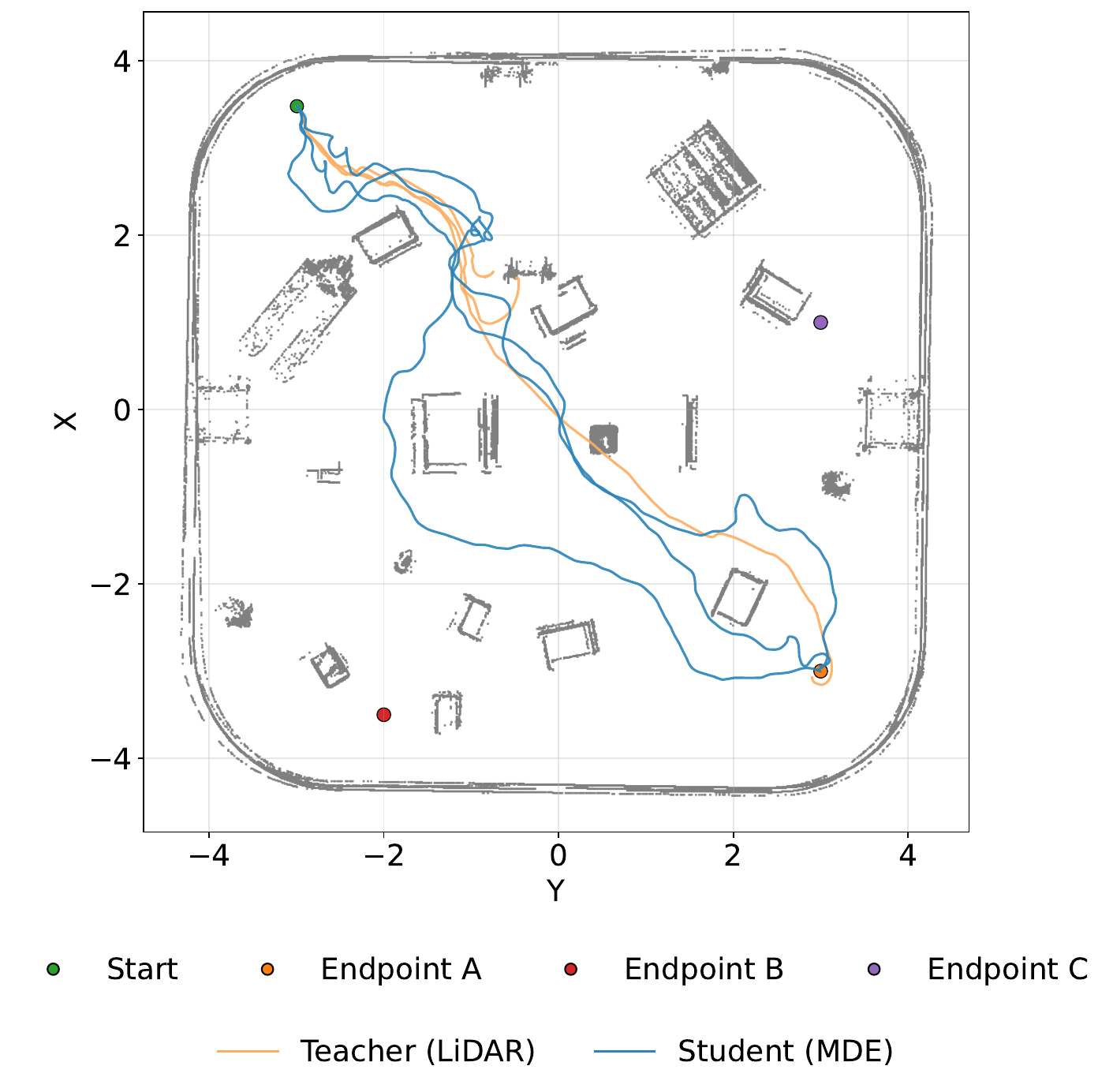}
    \caption{Visualization of the starting point and the three endpoints (axes in meters). Three trajectories for Endpoint~A are shown for both the teacher and student policies during real-world deployment.}
    \label{fig:student_2}
\end{figure}

\subsection{Sim-to-Real Analysis}

Fig.~\ref{fig:student_2} shows example trajectories for both policies during real-world deployment. Two of the three teacher runs collide with the ladder, while the student reaches the goal. However, the student trajectories exhibit oscillatory behavior. These oscillations are attributed to a combination of inference latency, residual \gls{mde} depth errors, and inaccuracies in the dynamic modeling of the real robot.

\begin{figure}[h!]
    \centering
        \includegraphics[width=1.0\columnwidth]{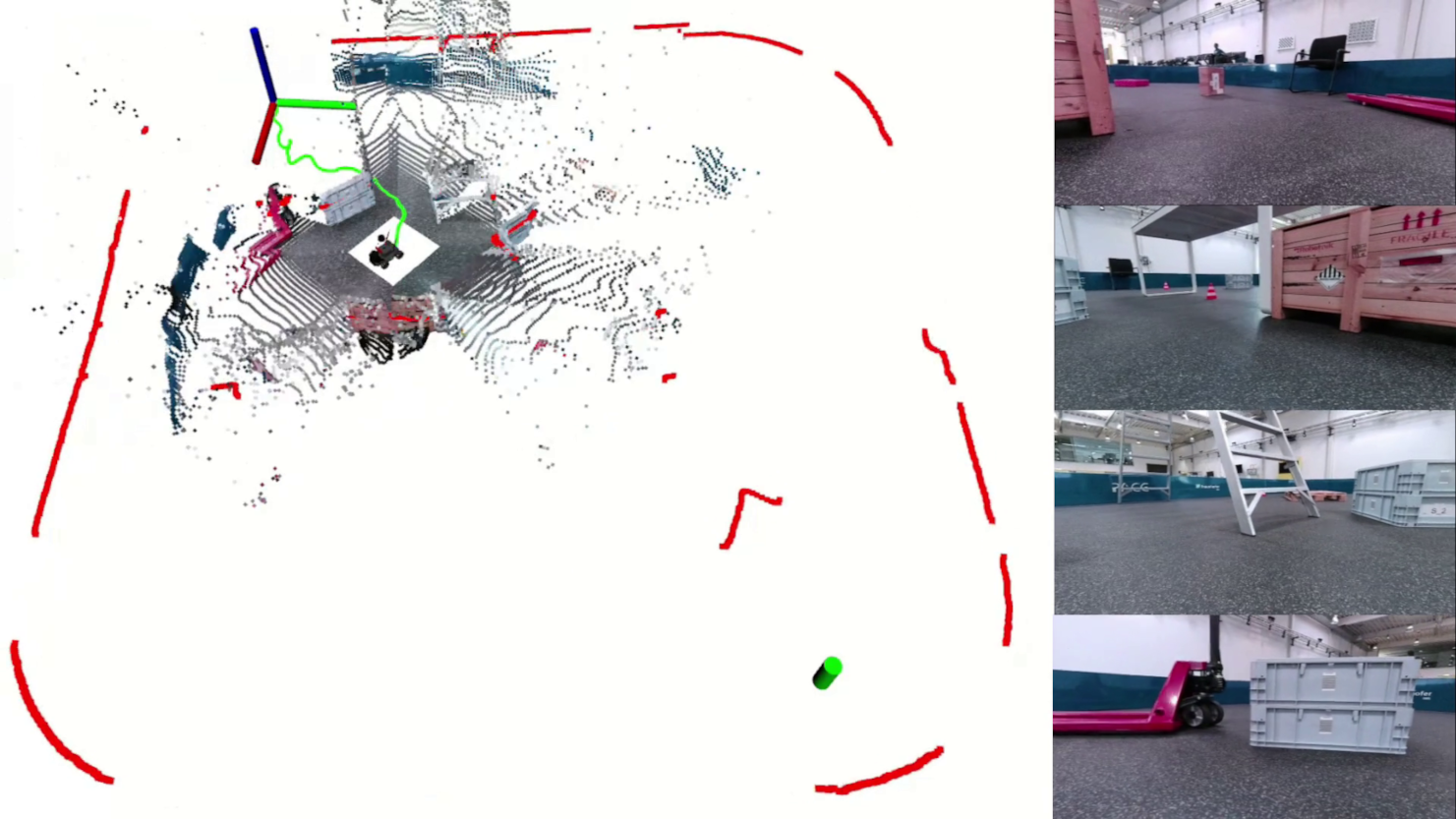}
    \caption{Point cloud comparison between ground-truth \gls{lidar} (red) and \gls{mde}-predicted depth. The scale mismatch increases at larger distances.}
    \label{fig:pcl_mismatch}
\end{figure}

Despite fine-tuning the Depth Anything V2 model on domain-specific data, a mismatch in absolute distance remains, as shown in Fig.~\ref{fig:pcl_mismatch}. For visualization, the figure displays a point cloud generated from the \gls{mde} depth maps using the camera calibration parameters from Section~\ref{sec:hardware_calibration}, alongside the absolute range measurements from the RPLIDAR S2 (shown in red). The \gls{mde} model estimates close-range distances with higher accuracy, while errors increase at larger distances, resulting in a residual sim-to-real gap.

\section{Conclusion and Outlook}

We presented a teacher-student framework that replaces 2D \gls{lidar} with \gls{mde} for vision-based mobile robot navigation. In simulation, the \gls{mde}-based student achieved success rates of 82--96.5\% across obstacle densities, consistently outperforming the standard 2D \gls{lidar} teacher (50--89\%). In real-world experiments, the student achieved an average success rate of 80\% compared to 37\% for the \gls{lidar}-based teacher, as the full-frame depth maps capture obstacles outside the \gls{lidar} scan plane. The complete inference pipeline runs onboard an NVIDIA Jetson Orin AGX at 10\,Hz without external computation for inference. Current limitations include the restriction to static environments, residual \gls{mde} scale errors at larger distances, and the absence of comparisons with external baselines, as differences in robot platforms and sensor configurations preclude a direct comparison. Future work will address dynamic obstacle avoidance and investigate online adaptation strategies to further close the sim-to-real gap.

\section*{ACKNOWLEDGMENT}
The authors acknowledge the use of Anthropic's Claude for language
refinement, including grammar, clarity, and conciseness improvements. All technical content, results, and interpretations
remain the sole work of the authors.

{\small
\setlength{\itemsep}{0pt}
\bibliographystyle{IEEEtran}
\bibliography{references}
}

\end{document}